\documentclass[11pt,letterpaper]{article}
\usepackage{emnlp2017}
\usepackage{times}
\usepackage{latexsym}
\usepackage{color}
\usepackage{paralist}
\usepackage{amsmath} 
\usepackage{amssymb}
\usepackage{xspace}
\usepackage{color}
\usepackage{graphicx}
\usepackage{enumitem}
\usepackage{booktabs}       
\usepackage{mysymbols}
\usepackage[normalem]{ulem}
\usepackage[font={footnotesize},labelfont={footnotesize}]{caption}
\usepackage{url}

\emnlpfinalcopy

\def\emnlppaperid{***}

\newcommand{\rama}[1]{\textcolor{black}{#1}}
\newcommand{\ramafinal}[1]{\textcolor{black}{#1}}
\newcommand{\ramacr}[1]{\textcolor{black}{#1}}
\newcommand{\ak}[1]{\textcolor{black}{#1}}

\newcommand{\final}[1]{\textcolor{black}{#1}}

\newcommand{\hjf}[1]{\textcolor{black}{#1}}
\newcommand{\cam}[1]{\textcolor{black}{#1}}

\newcommand{\refsec}[1]{Sec.~\ref{#1}}
\newcommand{\reftab}[1]{Table.~\ref{#1}}

\newcommand{\ebr}[1]{$\scriptstyle$\small{$\pm$#1}}

\title{Sound-Word2Vec: \\ Learning Word Representations Grounded in Sounds}

\author{Ashwin Vijayakumar\textsuperscript{1}, \hspace{1mm} Ramakrishna Vedantam\textsuperscript{2} \and Devi Parikh\textsuperscript{3,1} \\
    \textsuperscript{1} Georgia Tech, \textsuperscript{2} Virgina Tech, \textsuperscript{3} Facebook AI Research \\
\texttt{\{ashwinkv,parikh\}@gatech.edu, vrama1@vt.edu}}

\date{}

\begin{document}

\maketitle

\begin{abstract}
	To be able to interact better with humans, it is crucial for machines to understand sound -- a primary modality of human perception.
    Previous works have used sound to learn embeddings for improved generic semantic similarity assessment.
    In this work, we treat sound as a first-class citizen, studying \hjf{downstream} 6textual tasks which require aural grounding.
    To this end, we propose \emph{sound-word2vec} -- a new embedding scheme that learns \hjf{specialized word embeddings} grounded in sounds.
    For example, we learn that two seemingly (semantically) unrelated concepts, like \emph{leaves} and \emph{paper} are similar due to the similar \emph{rustling} sounds they make.
    Our embeddings prove useful in textual tasks requiring aural reasoning like text-based sound retrieval and discovering Foley sound effects (used in movies).
    Moreover, our embedding space captures interesting dependencies between words and onomatopoeia and outperforms prior work on aurally-relevant word relatedness datasets such as AMEN and ASLex.
\end{abstract}

\section{Introduction}
\vspace{-5pt}
Sound and vision are the dominant perceptual signals, while language helps us communicate complex experiences via rich abstractions.
For example, a novel 
can stimulate us to mentally construct the image of the scene despite having never physically perceived it.
Indeed, language has evolved to contain numerous constructs that help depict visual concepts.
For example, we can easily form the picture of a \emph{white, furry} cat with \emph{blue} eyes via. a description of the cat in terms of its visual attributes~\cite{Lampert,citeulike:10025772}.

\noindent\textbf{\emph{Need for Onomatopoeia.}} However, how would one describe the auditory instantiation of cats?
While a first thought might be to use 
audio descriptors like \emph{loud}, \emph{shrill}, \emph{husky} \etc as mid-level constructs or ``attributes'', arguably, it is difficult to precisely convey and comprehend 
sound through such language.
Indeed, \citet{wake_98} find that humans first communicate sounds using ``onomatopoeia'' -- words that \rama{are suggestive of the phonetics of} sounds while having no explicit meaning \eg \emph{meow, tic-toc}.
When asked for further explanation of sounds, humans provide descriptions of potential sound sources or impressions created by the sound (\emph{pleasant, annoying, \etc.})
\\ \\
\textbf{\emph{Need for Grounding in Sound.}} While onomatopoeic words exist for commonly found \rama{concepts}, 
a vast majority of concepts are not as perceptually striking or sufficiently frequent for us to come up with dedicated words describing their sounds.
Even worse, some sounds, say, musical instruments, might be difficult to mimic using speech.
Thus, for a large number of concepts there seems to be a gap between sound and its counterpart in language~\cite{narayan_aaai}.
This becomes problematic in specific situations where we want to talk about the heavy tail of concepts and their sounds, or while describing a particular sound we want to create as an effect (say in movies).
To alleviate this, a common literary strategy is to provide metaphors to more relatable exemplars.
For example, when we say, ``He thundered angrily'', we compare the person's angry speech to the sound of thunder to convey the seriousness of the situation.
However, without this grounding in sound, thunder and anger both appear to be seemingly unrelated concepts \rama{in terms of semantics}.
\\ \\
\textbf{\emph{Contributions.}} In this work, we learn embeddings to bridge the gap between sound and its counterpart in language.
\rama{We follow a retrofitting strategy, capturing similarity in sounds associated with words, \rama{while} using distributional semantics \rama{(from word2vec)} to provide smoothness to the embeddings. Note that we are not interested in capturing phonetic similarity, but the grounding in sound of the concept associated with the word (say ``rustling'' of leaves and paper.)}
We demonstrate the effectiveness of our embeddings on \rama{three} \hjf{downstream tasks that require reasoning about related aural cues}: \\
1.\hspace{3pt}Text-based sound retrieval -- Given a textual query describing the sound and a database containing sounds and associated textual tags, we retrieve \rama{sound samples by matching text} (\refsec{subsec:retrieval}) \\ 
2.\hspace{3pt}Foley Sound Discovery\hspace{-1pt} -- Given a short
phrase that outlines the technique of producing Foley sounds\footnote{Foley sounds are sound effects (typically ambient sounds) that are added to movies in the post-production stage to make actions or situations appear more realistic. These sounds are generally created using easily available proxy objects that mimic the sound of the true situation being depicted. For example, sound of breaking celery sticks is used to create the effect of breaking bones.}, we discover other relevant words (objects or actions) which can produce similar sound effects (\refsec{subsec:foley}) \\
\rama{3.\hspace{3pt}Aurally-relevant word relatedness assessment on AMEN and ASLex~\cite{kiela_emnlp15} (\refsec{subsec:amen})}
\\ \\
\noindent\rama{We also qualitatively compare with word2vec to highlight the unique notions of word relatedness captured by imposing auditory grounding.}

\label{sec:intro}
\section{Related Work}
\textbf{\emph{Audio and Word Embeddings.}} \hjf{Multiple works in the recent past \cite{bruni_jair14,lazaridou_arxiv15,lopopolo_iwcs2015,kiela_emnlp15,Kottur_2016_CVPR} have explored using perceptual modalities like vision and sound to learn language embeddings.
While \citet{lopopolo_iwcs2015} show preliminary results on using sound to learn distributional representations,
\citet{kiela_emnlp15} build on ideas from \citet{bruni_jair14} to learn word embeddings that respect both linguistic and auditory relationships by optimizing a joint objective.
Further, they propose various fusion strategies to combine knowledge from both the modalities. 
Instead, we ``specialize'' embeddings to exclusively \rama{respect} relationships defined by sounds, while initializing with \rama{word2vec embeddings for smoothness}.
Similar to previous findings \cite{melamud_arxiv16}, we observe that our specialized embeddings outperform language-only as well as other multi-modal embeddings in the downstream tasks of interest.}\\
\noindent\hjf{In an orthogonal and interesting direction,} other \hjf{recent works}~\cite{Chung_arxiv16,he_arxiv16,settle_arxiv16} learn \hjf{word representations based on similarity in their pronunciation and not the sounds associated with them}. In other words, phonetically similar words that have near identical pronunciations are brought closer in the embedding space \rama{(\eg, \emph{flower} and \emph{flour})}. \\
\noindent\citet{narayan_aaai}~\hjf{study the applicability of onomatopoeia to obtain semantically meaningful representations of audio. Using a novel word-similarity metric and principal component analysis,}
they find representations for sounds and cluster them in this derived space to reason about similarities.
In contrast, \hjf{we are interested in learning word representations that respect \rama{aural-}similarity.}
More importantly, our approach learns word representations for \ramafinal{in a data-driven manner} without having to \ramafinal{first} map the sound or its tags to \rama{corresponding} onomatopoeic words.
\\ \\
\textbf{\emph{Multimodal Learning with Surrogate Supervision.}} 
~\citet{Kottur_2016_CVPR} and \citet{Owens_ECCV} 
use a surrogate modality to induce supervision to learn representations \final{for a desired modality.} \rama{While \final{the former} learns word embeddings grounded in cartoon images, 
the \final{latter} learns visual features grounded in sound.} In contrast, we use sound as the surrogate modality to supervise representation learning for words.

\label{sec:related}
\section{Datasets}
\textbf{\emph{Freesound.}} We use the freesound database~\cite{freesound}, \rama{also used in prior work~\cite{kiela_emnlp15,lopopolo_iwcs2015}} to learn the proposed sound-word2vec embeddings. Freesound is a freely available, collaborative dataset consisting of user uploaded sounds permitting reuse.
All uploaded sounds have human descriptions in the form of tags and captions in natural language.
\ak{The tags contain a broad set of relevant topics for a sound (\eg, \emph{ambience, electronic, birds, city, reverb}) and captions describing the content of the sound, in addition to details pertaining to audio quality.}
For the text-based sound retrieval task, we use a subset of 234,120 sounds from this database and divide it into training (80\%), validation (10\%) and testing splits (10\%).
Further, for foley sound discovery, we aggregate descriptions of foley sound production provided by sound engineers \cite{epic_sound,foley_page} to create a list of 30 foley sound pairs, \ramafinal{which forms our ground truth for the task.}
\ak{For example, the description to produce a foley ``driving on gravel'' sound is to record the ``crunching sound of plastic or polyethene bags''.}
\\ \\
\noindent\textbf{\emph{AMEN and ASLex.}} AMEN and ASLex~\cite{kiela_emnlp15} are subsets of the standard MEN~\cite{bruni_jair14} and SimLex~\cite{hill_ACL_2015} word similarity datasets consisting of word-pairs that ``can be associated with a distinctive associated sound''.
We evaluate on this dataset for completeness to benchmark our approach against previous work. 
However, we are primarily interested in the slightly different problem of relating words with similar auditory instantions that may or may not be semantically related as opposed to relating semantically similar words that can be associated with some common auditory signal.

\label{sec:dataset}
\section{Approach}
\vspace{-5pt}
We use the Freesound database to construct a dataset of tuples $\{s, T\}$, where $s$ is a sound and $T$ is the set of associated user-provided tags.
We then aim to learn an embedding space for the tags that respects auditory grounding using sound information as cross-modal context -- similar to word2vec~\cite{mikolov_nips13} that uses neighboring words as context / supervision.
We now explain our approach in detail.
\\ \\
\noindent{\textbf{\textit{Audio Features and Clustering}.}} We represent each sound $s$ by a feature vector consisting of the mean and variance of the following audio descriptors that are readily available as part of Freesound database:
\begin{itemize}[leftmargin=*, noitemsep, nolistsep]
\item Mel-Frequency Cepstral Co-efficients: \ak{This feature represents the short-term power spectrum of an audio and closely approximates the response of the human auditory system -- computed as given in \cite{ganchev2005comparative}.}
\item Spectral Contrast: It is the magnitude difference in the peaks and valleys of the spectrum -- computed according to \cite{akkermans2009shape}.
\item Dissonance: It measures the perceptual roughness of the sound \cite{plomp1965tonal}.
\item Zero-crossing Rate: It is the percentage of sign changes between consecutive signal values and is indicative of noise content.
    \item Spectral Spread: This feature is the concatenation of the $k$-order moments of the spectrum, where $k \in \{0,1,2,3,4\}$.
    \item Pitch Salience: This feature helps discriminate between musical and non-musical tones. 
        While, pure tones and unpitched sounds have values near 0, musical sounds containing harmonics have higher values \cite{ricard2004towards}.
    \end{itemize}
We then use $K$-Means algorithm to cluster the sounds in this feature space to assign each sound to a cluster $C(s) \in \{1,\dots K\}$.
\cam{We set $K$ to 30 by evaluating the performance of the embeddings on text-based audio-retrieval on the held out validation set.
    Note that the clustering is only performed once, prior to representation learning described below.
}
\\ \\
\textbf{\emph{Representation Learning.}} We represent each tag $t \in T$ using a $|\calV|$ dimensional one-hot encoding denoted by $\mathbf{v}_t$, where $\calV$ is the set of all unique tags in the training set \ramacr{(the size of our dictionary)}.
This one-hot vector $\mathbf{v}_t$ is projected into a $D$-dimensional vector space via $W_{P} \in \mathbb{R}^{|\calV|\times D}$, the \emph{projection matrix}.
\ramacr{This projection matrix computes the representation for each word in $\calV$.
The idea of our approach is to use $W_{P}$ to accurately predict cluster assignments (for sounds associated with words), which enforces grounding in sound.}
For each data-point, we obtain the summary of the tags $T$, by averaging the projections of all tags in the set as $\frac{1}{|T|}\sum_{t \in T} W_{P}'\mathbf{v}_t$.
\begin{figure}[t]
    \includegraphics[width=1\columnwidth, height=4cm]{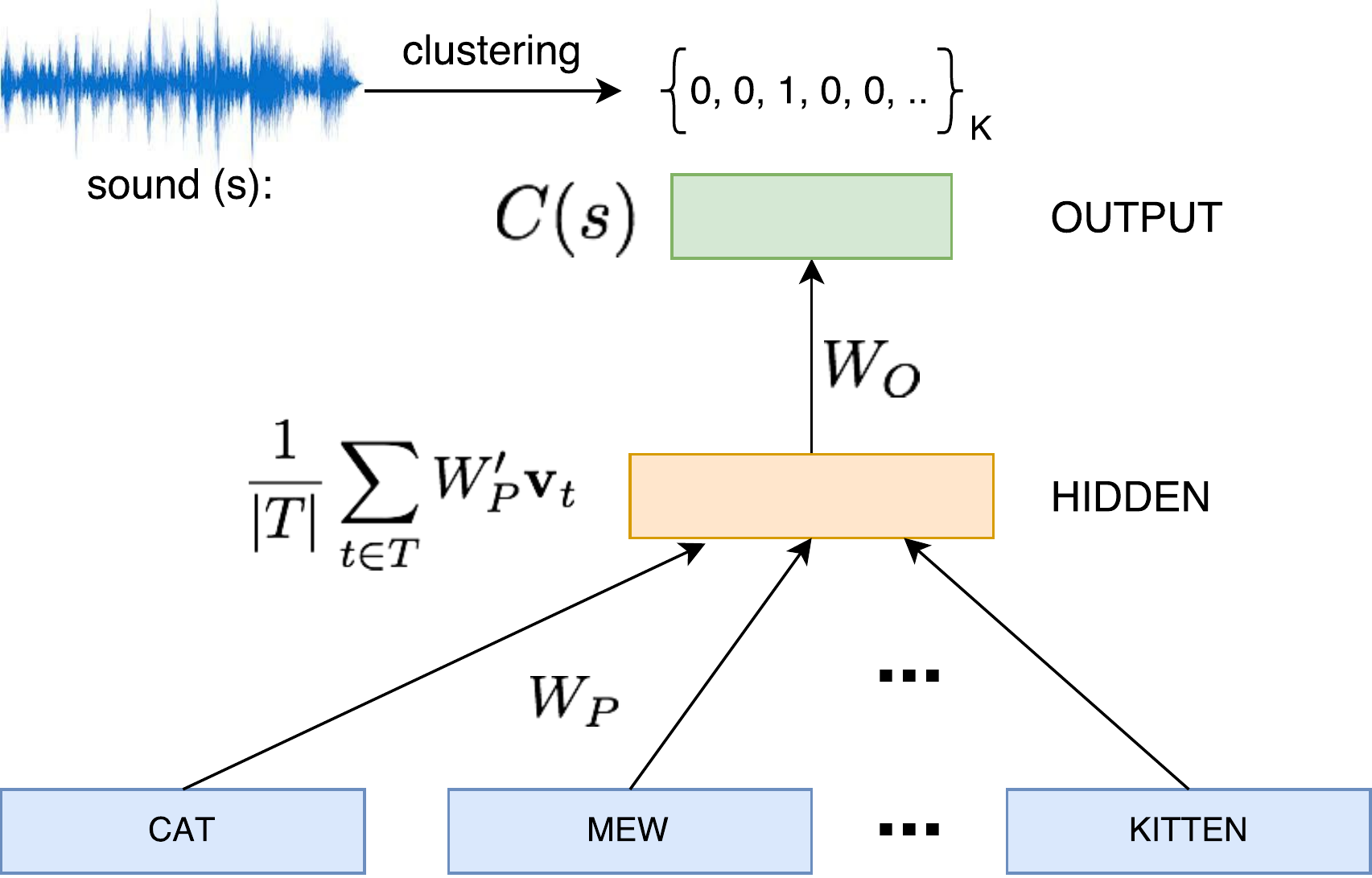}
    \caption{The model used to learn the proposed sound-word2vec embeddings. 
    The projection matrix $W_P$ containing that is used as the sound-word2vec embedding is learned by training the model to accurately predict the cluster assignment of the sound.}
    \vspace{-20pt}
\end{figure}
We then transform the so obtained summary representation via a linear layer (with parameters $W_O$) and pass the output through the softmax function to obtain a distribution, $p(c|T)$ over the K sound clusters.
We perform maximum-likelihood training for the correct cluster assignment $C(s)$\footnote{We also tried to regress directly to sound features instead of clustering, but found that it had poor performance.}, optimizing for parameters $W_P$ and $W_O$:
\begin{equation}
    \vspace{-10pt}
\max_{W_P, W_O} \log P(c=C(s)| T)
\end{equation}
We use SGD with momentum to optimize this objective which essentially is the cross-entropy between cluster assignments and $p(c| T)$.
We set $D$ to $300$ to be consistent with the publicly available word2vec embeddings. 
 \\ \\
\textbf{\emph{Initialization.}} We initialize $W_{P}$ with word2vec embeddings~\cite{mikolov_nips13} trained on the Google news corpus dataset with $\sim$3M words.
\ramafinal{We fine-tune on a subset of 9578 tags which are present in both Freesound as well as Google news corpus datasets, which is 55.68\% of the original tags in the Freesound dataset.
This helps us remove \cam{noisy} tags unrelated to the content of the sound.
}

In addition to enlarging the vocabulary, \ramafinal{the} pre-training helps induce smoothness in the sound-word2vec embeddings -- \ak{allowing us to 
\final{transfer} \ramafinal{semantics learnt from sounds to words that were not present as tags in the Freesound database}.}
\ramafinal{Indeed, we find that word2vec pre-training helps improve performance (\refsec{sec:results}).} Our use of language embeddings as an initialization to fine-tune \rama{(specialize)} from, as opposed to formulating a joint objective with language and audio context~\cite{kiela_emnlp15} is driven by the fact that we are interested in embeddings for words grounded in sounds, and not better generic word similarity.

\label{sec:approach}

\vspace{-5pt}
\section{Results}
\vspace{-5pt}
\textbf{\emph{Ablations}.} In addition to the language-only baseline word2vec \cite{mikolov_nips13}, we compare against tag-word2vec -- that predicts a tag using other tags of the sound as context, \rama{inspired by~\cite{font_ISWC_2014}. We also report results with a randomly initialized projection matrix (sound-word2vec(r)} to evaluate the effectiveness of pre-training with word2vec.

\noindent\ramafinal{\textbf{\emph{Prior work}.}} We compare against previous works \citet{lopopolo_iwcs2015} and \citet{kiela_emnlp15}. While the former uses a standard bag of words and SVD pipeline to arrive at distributional representations for words, the latter trains under a joint objective that respects both linguistic and auditory similarity. 
We use the openly available implementation for \citet{lopopolo_iwcs2015} and re-implement \citet{kiela_emnlp15} and train them on our dataset for a fair comparison of the methods.
In addition, we show a comparison to word-vectors released by~\cite{kiela_emnlp15} in the supplementary material. \rama{All approaches use an embedding size of 300 for consistency.}
\\
\\
\vspace{-30pt}
\subsection{Text-based Sound Retrieval}\label{subsec:retrieval}\label{subsec:retrieval}
Given a textual description of a sound as query, we compare it with tags associated with sounds in the database to retrieve the sound with the closest matching tags.
Note that this is a purely textual task, albeit one \rama{that needs awareness of} sound.
In a sense, this task exactly captures what we want our model to be able to do -- bridge the semantic gap between language and sound.
\rama{We use the training split (\refsec{sec:dataset}) to learn the sound-word2vec vectors, validation to pick the number of clusters (K), and report results on the test split.}
\rama{For retrieval, we represent sounds by averaging the learnt embeddings for the associated tags. We embed the caption provided for the sound (in the Freesound database) in a similar manner, and use it as the query. We then rank sounds based on the cosine similarity between the tag and query representations for retrieval.}
We evaluate using standard retrieval metrics -- \texttt{Recall@\{1,10,50,100\}}.
\cam{Note that the entire testing set ($\approx$10k sounds) is present in the retrieval pool.
    So, \texttt{recall@100} corresponds to obtaining the correct result in the top $1\%$ of the search results, which is a relatively stringent evaluation criterion.
}
\\ \\
\noindent\textbf{\emph{Results.}} \reftab{tab:ret} shows that our sound-word2vec embeddings outperform the baselines. \rama{We see that specializing the embeddings for sound using our two-stage training outperforms prior work(\citet{kiela_emnlp15} and~\citet{lopopolo_iwcs2015}), which did not do specialization.}
Among our approaches, tag-word2vec performs second best -- this is intuitive since the tag distributions \rama{implicitly capture auditory relatedness (a sound may have tags \emph{cat} and \emph{meow}), while word2vec and sound-word2vec(r) have the lowest performance.}
\begin{table}[t!]
    \centering
\resizebox{\columnwidth}{!}{
\begin{tabular}{ccccc} 
    \toprule
    Embedding & \multicolumn{4}{c}{\texttt{Recall}} \\
    & \texttt{@1} & \texttt{@10} & \texttt{@50} & \texttt{@100} \\
    \midrule
    word2vec & 6.47\ebr{0.00} & 14.25\ebr{0.05} & 21.72\ebr{0.12} & 26.03\ebr{0.22} \\
    tag-word2vec & 6.95\ebr{0.02}& 15.10\ebr{0.03} & 22.43\ebr{0.09}  & 27.21\ebr{0.24}\\
    sound-word2vec(r) & 6.49\ebr{0.00} & 14.98\ebr{0.03} & 21.96\ebr{0.11} & 26.43\ebr{0.20}\\
    \cite{lopopolo_iwcs2015}& 6.48\ebr{0.02}& 15.09\ebr{0.05} & 21.82\small\ebr{0.13} & 26.89\ebr{0.23}\\
    \cite{kiela_emnlp15}& 6.52\ebr{0.01}& 15.21\ebr{0.03} & 21.92\small\ebr{0.08} & 27.74\ebr{0.21}\\
    sound-word2vec & \textbf{7.11}\ebr{0.02} & \textbf{15.88}\ebr{0.04} & \textbf{23.14}\ebr{0.09} & \textbf{28.67}\ebr{0.17} \\
    \bottomrule
\end{tabular}
}
\vspace{-5pt}
\caption{Text-based sound retrieval (\emph{higher} is better). We find that our sound-word2vec model outperforms all baselines.}
\label{tab:ret}
\vspace{-5pt}
\end{table}
\vspace{-5pt}
\subsection{Foley Sound Discovery}\label{subsec:foley}
\rama{In this task, we evaluate how well embeddings identify matching pairs of target sounds (\emph{flapping bird wings}) and descriptions of Foley sound production techniques (\emph{rubbing a pair of gloves}).}
\rama{Intuitively, one expects sound-aware word embeddings to do better at this task than sound-agnostic ones.}
\final{We setup a ranking task by constructing a set \rama{of} original Foley sound pairs and decoy pairs formed by pairing the target description with every word from the vocabulary.}
\rama{We rank using cosine similarity \ramafinal{between the average word-vectors in each member of the pair}.}
\rama{A good embedding} is one in which the original Foley sound pair \rama{has the lowest rank.} We use the mean rank of the Foley sound in the dataset for evaluation.
\rama{We transfer the embeddings from~\refsec{subsec:retrieval} to this task, without additional training}.\\ \\
\noindent\textbf{\emph{Results.}}\hspace{3pt} We find that 	Sound-word2vec performs the best with a mean rank of 34.6 compared to other baselines tag-word2vec (38.9), \rama{sound-word2vec(r)} (114.3) and word2vec (189.45). \rama{\cam{As observed previously,} the second best performing approach is tag-word2vec.}
\hjf{\citet{lopopolo_iwcs2015} and \citet{kiela_emnlp15} perform worse than \rama{tag-word2vec} with a mean rank of 48.4 and 42.1 respectively.}
\rama{Note that random chance gets a} rank of $(|\calV|+1)/2 = 4789.5$.

\subsection{Evaluation on AMEN and ASLex}\label{subsec:amen}
\vspace{-5pt}
\begin{table}
    \centering
\resizebox{\columnwidth}{!}{
\begin{tabular}{ccc} 
    \toprule
    Embedding & \multicolumn{2}{c}{Spearman Correlation $\rho_s$} \\
    & AMEN & ASLex \\
    \midrule
    \cite{lopopolo_iwcs2015} & 0.410\ebr{0.09} & 0.237\ebr{0.04} \\
    \cite{kiela_emnlp15} & 0.648\ebr{0.08} & 0.366\ebr{0.11} \\
    sound-word2vec & \textbf{0.674}\ebr{0.05} & \textbf{0.391}\ebr{0.06} \\
    \bottomrule
\end{tabular}
}
\vspace{-5pt}
\caption{Comparison to state of the art AMEN and ASLex datasets~\cite{kiela_emnlp15} (\emph{higher} is better). \ramafinal{Our approach performs better than~\citet{kiela_emnlp15}.}}
\label{tab:amen}
\vspace{-15pt}
\end{table}
AMEN and ASLex~\cite{kiela_emnlp15} are subsets of the MEN and SimLex-999 datasets for word relatedness grounded in sound.
    \hjf{From \tabref{tab:amen}, we can see that our embeddings} outperform~\cite{kiela_emnlp15} \rama{on both AMEN and ASLex.}
    \rama{These datasets were curated by annotating concepts related by sound; however we observe that relatedness is often confounded}. \rama{For example, \emph{(river, water), (automobile, car)} are marked as aurally related — however they do not stand out as aurally-related examples as they are already semantically related.}
\ramafinal{In contrast, we are interested in how onomatopoeic words relate to regular words (\tabref{tab:2}), which we study by explicit grounding in sound. Thus while we show competitive performance on this dataset, it might not be best suited for studying the benefits of our approach.}

\label{sec:results}

\vspace{-5pt}
\section{Discussion and Conclusion}
\vspace{-5pt}

\begin{table}
    \centering
\resizebox{\columnwidth}{!}{
    \begin{tabular}{ccc}
    \toprule
    \textbf{word} & \textbf{word2vec} & \textbf{sound-word2vec} \\
    \midrule
    apple & apples, pear, fruit & bite, snack, chips \\
          & berry, pears, strawberry & chew, munch, carton \\
    \midrule
    wood & lumber, timber, softwoods, & wooden, snap, knock, \\
          & hardwoods, cedar, birch & smack, whack, snapping \\
    \midrule
    bones  & skull, femur, skeletons, & eggshell, carrot, arm\\
          & thighbone, pelvis, molar  & blood, polystyrene, crunch \\
    \midrule
    glass  & hand-blown, glassware, tumbler, & shattered, ceramic, smash \\
    & Plexiglass, wine-glass, bottle & clink, beer, spoon \\
    \midrule
    \textbf{Onomatopoeic query words} & & \\
    \midrule
    boom & booms, booming, bubble, &  bomb, bang, explosion\\
          & craze, downturn, upswing & bombing, exploding, ecstatic \\
    \midrule
    jingle & song, commercial, catchy-tune, &  magic, tinkle, nails\\
          & ditty, slogan, anthem & bells, key, doorbell \\
    \midrule
    slam & slams, piledriver, uranage &  shut, lock, opening\\
          & spinkick, hiptoss, hit & closing, latch, door \\
    \midrule
    quack & charlatan, quackery, crackpot &  duck, snort, calling\\
          & homeopaths, concoctions, snake-oil & chirp, tweet, oink \\
    \bottomrule
\end{tabular}}
\vspace{-5pt}
\caption{We show nearest neighbors in both word2vec and sound-word2vec spaces for eight words (`regular' words, top half and onomatopoeic words, bottom half).}
\label{tab:2}
\vspace{-15pt}
\end{table}
We show nearest neighbors in both sound-word2vec and word2vec space (\tabref{tab:2}) to qualitatively demonstrate the unique dependencies captured due to auditory grounding.
While word2vec maps a word (say, \emph{apple}) to other semantically similar words (other fruits), similar `sounding' words (\emph{chips}) or onomatopoeia (\emph{munch}) are closer in our embedding space.
Moreover, onomatopoeic words (say, \emph{boom} and \emph{slam}) are mapped to relevant objects (\emph{explosion} and \emph{door}).
Interestingly, parts (\eg, \emph{lock, latch}) and actions (\emph{closing}) are also closer to the onomatopoeic query -- exhibiting an understanding of the auditory scene. \\
\noindent\rama{\textbf{\emph{Conclusion.}} In this work we introduce a novel word embedding scheme that respects auditory grounding. We show that our embeddings provide strong performance on text-based sound retrieval, Foley sound discovery along with intuitive nearest neighbors for onomatopoeia that are tasks in text requiting auditory reasoning. 
We hope our work motivates further efforts on understanding and relating onomatopoeia words to ``regular'' words.}
\vspace{5pt}

\noindent{\tiny \textbf{\emph{Acknowledgements. }}
    We thank the Freesound team and Frederic Font in particular for helping us with the Freesound API. We also thank Khushi Gupta and Stefan Lee for fruitful discussions. This work was funded in part by an NSF CAREER, ONR Grant N00014-16-1-2713, ONR YIP, Sloan Fellowship, Allen Distinguished Investigator, Google Faculty Research Award, Amazon Academic Research Award to DP. \par
}


\label{sec:discussion}

\bibliography{ashwinkv}
\bibliographystyle{emnlp_natbib}

\end{document}